\documentclass[11pt,a4paper]{article}
\usepackage[hyperref]{emnlp2018}
\usepackage{times}
\usepackage{latexsym}

\usepackage{url}

\aclfinalcopy 

\title{Adapting Deep Learning Methods for\\Mental Health Prediction on Social Media}

\author{Ivan Sekuli\'c and Michael Strube\\
  Heidelberg Institute for Theoretical Studies gGmbH \\
  \tt \{ivan.sekulic,michael.strube\}@h-its.org
  }
  
\date{}

\usepackage[group-separator={,}]{siunitx}

\usepackage[inline]{enumitem}
\usepackage{booktabs}
\usepackage{graphicx}
\begin{document}
\maketitle
\begin{abstract}
  Mental health poses a significant challenge for an individual's well-being. Text analysis of rich resources, like social media, can contribute to deeper understanding of illnesses and provide means for their early detection. We tackle a challenge of detecting social media users' mental status through deep learning-based models, moving away from traditional approaches to the task. In a binary classification task on predicting if a user suffers from one of nine different disorders, a hierarchical attention network outperforms previously set benchmarks for four of the disorders. Furthermore, we explore the limitations of our model and analyze phrases relevant for classification by inspecting the model's word-level attention weights.

  
\end{abstract}

\section{Introduction}

Mental health is a serious issue of the modern-day world.
According to the World Health Organization's 2017 report and \newcite{wykes2015mental} more than a quarter of Europe's adult population suffers from an episode of a mental disorder in their life.
The problem grows bigger with the fact that as much as 35--50\% of those affected go undiagnosed and receive no treatment for their illness.
In line with WHO's Mental Health Action Plan \cite{saxena2013world}, the natural language processing community helps the gathering of information and evidence on mental conditions, focusing on text analysis of authors affected by mental illnesses.


Researchers can utilize large amounts of text on social media sites to get a deeper understanding of mental health and develop models for early detection of various mental disorders \cite{de2013social, coppersmith2014quantifying, gkotsis2016language, benton2017multi, sekulic2018not, zomick2019linguistic}.
In this work, we experiment with the Self-reported Mental Health Diagnoses (SMHD) dataset \cite{cohan2018smhd}, consisting of thousands of Reddit users diagnosed with one or more mental illnesses.
The contribution of our work is threefold. 
First, we adapt a deep neural model, proved to be successful in large-scale document classification, for user classification on social media, outperforming previously set benchmarks for four out of nine disorders. In contrast to the majority of preceding studies on mental health prediction in social media, which relied mostly on traditional classifiers, we employ Hierarchical Attention Network (HAN) \cite{yang2016hierarchical}.
Second, we explore the limitations of the model in terms of data needed for successful classification, specifically, the number of users and number of posts per user.
Third, through the attention mechanism of the model, we analyze the most relevant phrases for the classification and compare them to previous work in the field.
We find similarities between lexical features and n-grams identified by the attention mechanism, supporting previous analyses.



\section{Dataset and the Model}
\label{dataset}

\subsection{Self-reported Mental Health Diagnoses Dataset}
The SMHD dataset \cite{cohan2018smhd} is a large-scale dataset of Reddit posts from users with one or multiple mental health conditions.
The users were identified by constructing patterns for discovering self-reported diagnoses of nine different mental disorders.
For example, if a user writes \emph{``I was officially diagnosed with depression last year''}, she/he/other would be considered to suffer from depression.

Nine or more control users, which are meant to represent general population, are selected for each diagnosed user by their similarity, i.e., by their number of posts and the subreddits (sub-forums on Reddit) they post in.
Diagnosed users' language is normalized by removing posts with specific mental health signals and discussions, in order to analyze the language of general discussions and to be more comparable to the control groups.
The nine disorders and the number of users per disorder, as well as average number of posts per user, are shown in Table~\ref{tbl:desc}.

\begin{table}[]
\centering

\begin{tabular}{lll}
\toprule
& \multicolumn{1}{c}{\# users} & \multicolumn{1}{c}{\begin{tabular}[c]{@{}c@{}}\# posts\\ per user\end{tabular}}\\
\midrule
Depression    & 14,139                       & 162.2 (84.2)                                                                    \\
ADHD          & 10,098                       & 164.7 (83.6)                                                                    \\
Anxiety       & 8,783                        & 159.7 (83.0)                                                                    \\
Bipolar       & 6,434                        & 157.6 (82.4)                                                                    \\
PTSD          & 2,894                        & 160.7 (84.7)                                                                    \\
Autism        & 2,911                        & 168.3 (84.5)                                                                    \\
OCD           & 2,336                        & 158.8 (81.4)                                                                    \\
Schizophrenia & 1,331                        & 157.3 (80.5)                                                                    \\
Eating        & 598                          & 161.4 (81.0)\\                                                  
\bottomrule
\end{tabular}
\caption{\label{tbl:desc} Number of users in SMHD dataset per condition and the average number of posts per user (with std.).}

\end{table}

For each disorder, \newcite{cohan2018smhd} analyze the differences in language use between diagnosed users and their respective control groups.
They also provide benchmark results for the binary classification task of predicting whether the user belongs to the diagnosed or the control group.
We reproduce their baseline models for each disorder and compare to our deep learning-based model, explained in Section~\ref{han_explained}. 

\subsection{Selecting the Control Group}
\label{control}
\newcite{cohan2018smhd} select nine or more control users for each diagnosed user and run their experiments with these mappings. 
With this exact mapping not being available, for each of the nine conditions, we had to select the control group ourselves.
For each diagnosed user, we draw exactly nine control users from the pool of \num[group-separator={,}]{335952} control users present in SMHD and proceed to train and test our binary classifiers on the newly created sub-datasets.

In order to create a statistically-fair comparison, we run the selection process multiple times, as well as reimplement the benchmark models used in \newcite{cohan2018smhd}.
Multiple sub-datasets with different control groups not only provide us with unbiased results, but also show how results of a binary classification can differ depending on the control group.

\subsection{Hierarchical Attention Network}
\label{han_explained}
We adapt a Hierarchical Attention Network (HAN) \cite{yang2016hierarchical}, originally used for document classification, to user classification on social media.
A HAN consists of a word sequence encoder, a word-level attention layer, a sentence encoder and a sentence-level attention layer.
It employs GRU-based sequence encoders \cite{cho2014properties} on sentence and document level, yielding a document representation in the end.
The word sequence encoder produces a representation of a given sentence, which then is forwarded to a sentence sequence encoder that, given a sequence of encoded sentences, returns a document representation.
Both, word sequence and sentence sequence encoders, apply attention mechanisms on top to help the encoder more accurately aggregate the representation of given sequence.
For details of the architecture we refer the interested readers to \newcite{yang2016hierarchical}.

In this work, we model a user as a document, enabling an intuitive adaptation of the HAN.
Just as a document is a sequence of sentences, we propose to model a social media user as a sequence of posts.
Similarly, we identify posts as sentences, both being a sequence of tokens.
This interpretation enables us to apply the HAN, which had great success in document classification, to user classification on social media. 

\section{Results}
\label{results}
\begin{table*}[t!]
\centering
\resizebox{\textwidth}{!}{
\begin{tabular}{lllllllllll}
\toprule
                    & Depression & ADHD  & Anxiety & Bipolar & PTSD  & Autism & OCD   & Schizo & Eating &  \\
                    \midrule
Logistic Regression & 59.00      & 51.02 & 62.34   & 61.87   & 69.34 & \textbf{55.57}  & \textbf{59.49} & 56.31         & 70.71  &  \\
Linear SVM          & 58.64      & 50.08 & 61.69   & 61.30   & \textbf{69.91} & 55.35  & 58.56 & \textbf{57.43}         & \textbf{70.91} &  \\
Supervised FastText & 58.38      & 48.80 & 60.17   & 56.53   & 61.08 & 49.52  & 54.16 & 46.73         & 63.73  &  \\
HAN                 & \textbf{68.28} & \textbf{64.27} & \textbf{69.24} & \textbf{67.42}   & 68.59 & 53.09  & 58.51 & 53.68         & 63.94  &  \\
\bottomrule
\end{tabular}
}
\caption{\label{tbl:rez} $F_1$ measure averaged over five runs with different control groups.}
\end{table*}

\subsection{Experimental Setup}
The HAN uses two layers of bidirectional GRU units with hidden size of 150, each of them followed by a 100 dimensional attention mechanism. 
The first layer encodes posts, while the second one encodes a user as a sequence of encoded posts. 
The output layer is 50-dimensional fully-connected network, with binary cross entropy as a loss function.
We initialize the input layer with 300 dimensional GloVe word embeddings \cite{pennington2014glove}.
We train the model with Adam \cite{kingma2014adam}, with an initial learning rate of $10^{-4}$ and a batch size of 32 for 50 epochs. 
The model that proves best on the development set is selected. 

We implement the baselines as in \newcite{cohan2018smhd}.
Logistic regression and the linear SVM were trained on \emph{tf-idf} weighted bag-of-words features, where users' posts are all concatenated and all the tokens lower-cased. 
Optimal parameters were found on the development set, and models were evaluated on the test set.
FastText \cite{joulin2016fasttext} was trained for 100 epochs, using character n-grams of size 3 to 6, with a 100 dimensional hidden layer. 
We take diagnosed users from predefined train-dev-test split, and select the control group as described in Subsection~\ref{control}.
To ensure unbiased results and fair comparison to the baselines, we repeat the process of selecting the control group five times for each disorder and report the average of the runs.

\subsection{Binary Classification per Disorder}
We report the $F_1$ measures per disorder in Table~\ref{tbl:rez}, in the task of binary classification of users, with the diagnosed class as a positive one. 
Our model outperforms the baseline models for \emph{Depression}, \emph{ADHD}, \emph{Anxiety}, and \emph{Bipolar} disorder, while it proves insufficient for \emph{PTSD}, \emph{Autism}, \emph{OCD}, \emph{Schizophrenia}, and \emph{Eating} disorder.
We hypothesize that the reason for this are the sizes of particular sub-datasets, which can be seen in Table~\ref{tbl:desc}.
We observe higher $F_1$ score for the HAN in disorders with sufficient data, suggesting once again that deep neural models are data-hungry \cite{sun2017revisiting}. 
Logistic regression and linear SVM achieve higher scores where there is a smaller number of diagnosed users.
In contrast to \newcite{cohan2018smhd}, supervised FastText yields worse results than tuned linear models.

\begin{figure}[]
    \centering
    \includegraphics[scale=0.5]{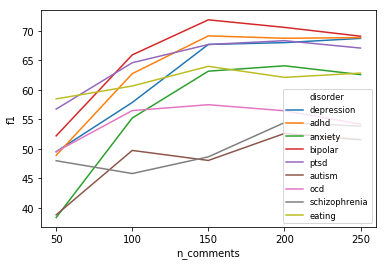}
    \caption{$F_1$ scores for different number of posts per users made available to HAN, averaged over three runs for different control groups.}
    \label{fig:my_label}
\end{figure}

We further investigate the impact of the size of the dataset on the final results of classification.
We limit the number of posts per user available to the model to examine the amount needed for reasonable performance.
The results for $50$, $100$, $150$, $200$, and $250$ posts per user available are presented in Figure~\ref{fig:my_label}.
Experiments were run three times for each disorder and each number of available posts, every time with a different control group selected.
We observe a positive correlation between the data provided to the model and the performance, although we find an upper bound to this tendency.
As the average number of posts per user is roughly $160$ (Table~\ref{tbl:desc}), it is reasonable to expect of a model to perform well with similar amounts of data available.
However, further analysis is required to see if the model reaches the plateau because a large amount of data is not needed for the task, or due to it not being expressive enough.


\subsection{Attention Weights Analysis}

The HAN, through attention mechanism, provides a clear way to identify posts, and words or phrases in those posts, relevant for classification.
We examine attention weights on a word level and compare the most attended words to prior research on depression.
Depression is selected as the most prevalent disorder in the SMHD dataset with a number of studies in the field \cite{rude2004language, chung2007psychological, de2013predicting, park2012depressive}.
For each post, we extracted two words with the highest attention weight as being the most relevant for the classification.
If the two words are appearing next to each other in a post we consider them as bigram.
Some of the top $100$ most common unigrams and bigrams are presented in Table~\ref{tbl:attention}, aggregated under the most common LIWC categories.

We observe similar patterns in features shown relevant by the HAN and previous research on signals of depression in language.
The importance of personal pronouns in distinguishing depressed authors from the control group is supported by multiple studies \cite{rude2004language, chung2007psychological, de2013predicting, cohan2018smhd}.
In the categories \textit{Affective processes}, \textit{Social processes}, and \textit{Biological processes}, \newcite{cohan2018smhd} report significant differences between depressed and control group, similar to some other disorders.
Except the above mentioned words and their abbreviations, among most commonly attended are swear words, as well as other forms of informal language.
The attention mechanism's weighting suggests that words and phrases proved important in previous studies, using lexical features and linear models, are relevant for the HAN as well.

\begin{table}[]

\centering
\resizebox{\columnwidth}{!}{
\begin{tabular}{lll}
\toprule
                     & unigrams                                & bigrams                                      \\
                     \midrule
Pers.\,pronouns    & \textit{I, my, her, your, they}         & \textit{I've never, your thoughts} \\
Affective  & \textit{like, nice, love, bad}          & \textit{I love}                              \\
Social     & \textit{friend, boyfriend, girl, guy}   & \textit{my dad, my girlfriend, my ex}        \\
Biological & \textit{pain, sex, skin, sleep, porn}   & \textit{your pain, a doctor, a therapist}    \\
Informal    & \textit{omg, lol, shit, fuck, cool}     & \textit{tl dr, holy shit}                    \\
Other                & \textit{advice, please, reddit} & \textit{thank you, your advice}             \\\bottomrule
\end{tabular}}
\caption{\label{tbl:attention} Unigrams and bigrams most often given the highest weight by attention mechanism in depression classification.}
\end{table}

\section{Related Work}
In recent years, social media has been a valuable source for psychological research.
While most studies use Twitter data \cite{coppersmith2015adhd, coppersmith2014quantifying, benton2017multi, coppersmith2015clpsych}, a recent stream turns to Reddit as a richer source of high-volume data \cite{de2014mental, shen2017detecting, gjurkovic2018reddit, cohan2018smhd, sekulic2018not, zirikly-etal-2019-clpsych}.
Previous approaches to author's mental health prediction usually relied on linguistic and stylistic features, e.g., Linguistic Inquiry and Word Count (LIWC) \citep{pennebaker2001linguistic} -- a widely used feature extractor for various studies regarding mental health \cite{rude2004language, coppersmith2014quantifying, sekulic2018not, zomick2019linguistic}.

Recently, \newcite{song2018feature} built a feature attention network for depression detection on Reddit, showing high interpretability, but low improvement in accuracy.
\newcite{orabi2018deep} concatenate all the tweets of a Twitter user in a single document and experiment with various deep neural models for depression detection.
Some of the previous studies use deep learning methods on a post level to infer general information about a user \cite{kshirsagar2017detecting, ive2018hierarchical, ruder2016character}, or detect different mental health concepts in posts themselves \cite{rojas2018deep}, while we focus on utilizing all of the users' text.
\newcite{yates2017depression} use a CNN on a post-level to extract features, which are then concatenated to get a user representation used for self-harm and depression assessment. 
A CNN requires a fixed length of posts, putting constraints on the data available to the model, while a HAN utilizes all of the data from posts of arbitrary lengths.


A social media user can be modeled as collection of their posts, so we look at neural models for large-scale text classification.
\newcite{liu2018long} split a document into chunks and use a combination of CNNs and RNNs for document classification. While this approach proves to be successful for scientific paper categorization, it is unintuitive to use in social media text due to an unclear way of splitting user's data into equally sized chunks of text.
\newcite{yang2016hierarchical} use a hierarchical attention network for document classification, an approach that we adapt for Reddit.
A step further would be adding another hierarchy, similar to \newcite{jiang2019semantic}, who use a multi-depth attention-based hierarchical RNN to tackle the problem of long-length document semantic matching.

\subsection{Ethical considerations}
Acknowledging the social impact of NLP research \cite{hovy2016social}, mental health analysis must be approached carefully as it is an extremely sensitive matter \cite{vsuster2017short}.
In order to acquire the SMHD dataset, we comply to the Data Usage Agreement, made to protect the users' privacy.
We do not attempt to contact the users in the dataset, nor identify or link them with other user information.

\section{Conclusion}
\label{conclusion}
In this study, we experimented with hierarchical attention networks for the task of predicting mental health status of Reddit users.
For the disorders with a fair amount of diagnosed users, a HAN proves to be better than the baselines.
However, the results worsen as the data available decreases, suggesting that traditional approaches remain better for smaller datasets.
The analysis of attention weights on word level suggested similarities to previous studies of depressed authors.
Embedding mental health-specific insights from previous work could benefit the model in general.
Future work includes analysis of post-level attention weights, with a goal of finding patterns in the relevance of particular posts, and, through them, time periods when a user is in distress.
As some of the disorders share similar symptoms, e.g., depressive episodes in bipolar disorder, exploiting correlations between labels through multi-task or transfer learning techniques might prove useful. 
In order to improve the classification accuracy, a transformer-based model for encoding users' posts should be tested.

\section{Acknowledgments}
This work has been funded by the Erasmus+ programme of the European Union and the Klaus Tschira Foundation.

\bibliography{emnlp2018}

\begin{thebibliography}{37}
\expandafter\ifx\csname natexlab\endcsname\relax\def\natexlab#1{#1}\fi

\bibitem[{Benton et~al.(2017)Benton, Mitchell, and Hovy}]{benton2017multi}
Adrian Benton, Margaret Mitchell, and Dirk Hovy. 2017.
\newblock Multi-task learning for mental health using social media text.
\newblock \emph{arXiv preprint arXiv:1712.03538}.

\bibitem[{Cho et~al.(2014)Cho, Van~Merri{\"e}nboer, Bahdanau, and
  Bengio}]{cho2014properties}
Kyunghyun Cho, Bart Van~Merri{\"e}nboer, Dzmitry Bahdanau, and Yoshua Bengio.
  2014.
\newblock On the properties of neural machine translation: Encoder-decoder
  approaches.
\newblock \emph{arXiv preprint arXiv:1409.1259}.

\bibitem[{Chung and Pennebaker(2007)}]{chung2007psychological}
Cindy Chung and James~W Pennebaker. 2007.
\newblock The psychological functions of function words.
\newblock \emph{Social communication}, 1:343--359.

\bibitem[{Cohan et~al.(2018)Cohan, Desmet, Yates, Soldaini, MacAvaney, and
  Goharian}]{cohan2018smhd}
Arman Cohan, Bart Desmet, Andrew Yates, Luca Soldaini, Sean MacAvaney, and
  Nazli Goharian. 2018.
\newblock Smhd: a large-scale resource for exploring online language usage for
  multiple mental health conditions.
\newblock In \emph{27th International Conference on Computational Linguistics},
  pages 1485--1497. ACL.

\bibitem[{Coppersmith et~al.(2014)Coppersmith, Dredze, and
  Harman}]{coppersmith2014quantifying}
Glen Coppersmith, Mark Dredze, and Craig Harman. 2014.
\newblock Quantifying mental health signals in twitter.
\newblock In \emph{Proceedings of the workshop on computational linguistics and
  clinical psychology: From linguistic signal to clinical reality}, pages
  51--60.

\bibitem[{Coppersmith et~al.(2015{\natexlab{a}})Coppersmith, Dredze, Harman,
  and Hollingshead}]{coppersmith2015adhd}
Glen Coppersmith, Mark Dredze, Craig Harman, and Kristy Hollingshead.
  2015{\natexlab{a}}.
\newblock From adhd to sad: Analyzing the language of mental health on twitter
  through self-reported diagnoses.
\newblock In \emph{Proceedings of the 2nd Workshop on Computational Linguistics
  and Clinical Psychology: From Linguistic Signal to Clinical Reality}, pages
  1--10.

\bibitem[{Coppersmith et~al.(2015{\natexlab{b}})Coppersmith, Dredze, Harman,
  Hollingshead, and Mitchell}]{coppersmith2015clpsych}
Glen Coppersmith, Mark Dredze, Craig Harman, Kristy Hollingshead, and Margaret
  Mitchell. 2015{\natexlab{b}}.
\newblock Clpsych 2015 shared task: Depression and ptsd on twitter.
\newblock In \emph{Proceedings of the 2nd Workshop on Computational Linguistics
  and Clinical Psychology: From Linguistic Signal to Clinical Reality}, pages
  31--39.

\bibitem[{De~Choudhury et~al.(2013{\natexlab{a}})De~Choudhury, Counts, and
  Horvitz}]{de2013social}
Munmun De~Choudhury, Scott Counts, and Eric Horvitz. 2013{\natexlab{a}}.
\newblock Social media as a measurement tool of depression in populations.
\newblock In \emph{Proceedings of the 5th Annual ACM Web Science Conference},
  pages 47--56. ACM.

\bibitem[{De~Choudhury and De(2014)}]{de2014mental}
Munmun De~Choudhury and Sushovan De. 2014.
\newblock Mental health discourse on reddit: Self-disclosure, social support,
  and anonymity.
\newblock In \emph{Eighth International AAAI Conference on Weblogs and Social
  Media}.

\bibitem[{De~Choudhury et~al.(2013{\natexlab{b}})De~Choudhury, Gamon, Counts,
  and Horvitz}]{de2013predicting}
Munmun De~Choudhury, Michael Gamon, Scott Counts, and Eric Horvitz.
  2013{\natexlab{b}}.
\newblock Predicting depression via social media.
\newblock In \emph{Seventh international AAAI conference on weblogs and social
  media}.

\bibitem[{Gjurkovi{\'c} and {\v{S}}najder(2018)}]{gjurkovic2018reddit}
Matej Gjurkovi{\'c} and Jan {\v{S}}najder. 2018.
\newblock Reddit: A gold mine for personality prediction.
\newblock In \emph{Proceedings of the Second Workshop on Computational Modeling
  of People’s Opinions, Personality, and Emotions in Social Media}, pages
  87--97.

\bibitem[{Gkotsis et~al.(2016)Gkotsis, Oellrich, Hubbard, Dobson, Liakata,
  Velupillai, and Dutta}]{gkotsis2016language}
George Gkotsis, Anika Oellrich, Tim Hubbard, Richard Dobson, Maria Liakata,
  Sumithra Velupillai, and Rina Dutta. 2016.
\newblock The language of mental health problems in social media.
\newblock In \emph{Proceedings of the Third Workshop on Computational
  Linguistics and Clinical Psychology}, pages 63--73.

\bibitem[{Hovy and Spruit(2016)}]{hovy2016social}
Dirk Hovy and Shannon~L Spruit. 2016.
\newblock The social impact of natural language processing.
\newblock In \emph{Proceedings of the 54th Annual Meeting of the Association
  for Computational Linguistics (Volume 2: Short Papers)}, pages 591--598.

\bibitem[{Ive et~al.(2018)Ive, Gkotsis, Dutta, Stewart, and
  Velupillai}]{ive2018hierarchical}
Julia Ive, George Gkotsis, Rina Dutta, Robert Stewart, and Sumithra Velupillai.
  2018.
\newblock Hierarchical neural model with attention mechanisms for the
  classification of social media text related to mental health.
\newblock In \emph{Proceedings of the Fifth Workshop on Computational
  Linguistics and Clinical Psychology: From Keyboard to Clinic}, pages 69--77.

\bibitem[{Jiang et~al.(2019)Jiang, Zhang, Li, Bendersky, Golbandi, and
  Najork}]{jiang2019semantic}
Jyun-Yu Jiang, Mingyang Zhang, Cheng Li, Michael Bendersky, Nadav Golbandi, and
  Marc Najork. 2019.
\newblock Semantic text matching for long-form documents.
\newblock In \emph{The World Wide Web Conference}, pages 795--806. ACM.

\bibitem[{Joulin et~al.(2016)Joulin, Grave, Bojanowski, Douze, J{\'e}gou, and
  Mikolov}]{joulin2016fasttext}
Armand Joulin, Edouard Grave, Piotr Bojanowski, Matthijs Douze, H{\'e}rve
  J{\'e}gou, and Tomas Mikolov. 2016.
\newblock Fasttext.zip: Compressing text classification models.
\newblock \emph{arXiv preprint arXiv:1612.03651}.

\bibitem[{Kingma and Ba(2014)}]{kingma2014adam}
Diederik~P Kingma and Jimmy Ba. 2014.
\newblock Adam: A method for stochastic optimization.
\newblock \emph{arXiv preprint arXiv:1412.6980}.

\bibitem[{Kshirsagar et~al.(2017)Kshirsagar, Morris, and
  Bowman}]{kshirsagar2017detecting}
Rohan Kshirsagar, Robert Morris, and Sam Bowman. 2017.
\newblock Detecting and explaining crisis.
\newblock \emph{arXiv preprint arXiv:1705.09585}.

\bibitem[{Liu et~al.(2018)Liu, Liu, Cong, Zhao, Ji, and He}]{liu2018long}
Liu Liu, Kaile Liu, Zhenghai Cong, Jiali Zhao, Yefei Ji, and Jun He. 2018.
\newblock Long length document classification by local convolutional feature
  aggregation.
\newblock \emph{Algorithms}, 11(8):109.

\bibitem[{Orabi et~al.(2018)Orabi, Buddhitha, Orabi, and
  Inkpen}]{orabi2018deep}
Ahmed~Husseini Orabi, Prasadith Buddhitha, Mahmoud~Husseini Orabi, and Diana
  Inkpen. 2018.
\newblock Deep learning for depression detection of twitter users.
\newblock In \emph{Proceedings of the Fifth Workshop on Computational
  Linguistics and Clinical Psychology: From Keyboard to Clinic}, pages 88--97.

\bibitem[{Park et~al.(2012)Park, Cha, and Cha}]{park2012depressive}
Minsu Park, Chiyoung Cha, and Meeyoung Cha. 2012.
\newblock Depressive moods of users portrayed in twitter.
\newblock In \emph{Proceedings of the ACM SIGKDD Workshop on healthcare
  informatics (HI-KDD)}, volume 2012, pages 1--8.

\bibitem[{Pennebaker et~al.(2001)Pennebaker, Francis, and
  Booth}]{pennebaker2001linguistic}
James~W Pennebaker, Martha~E Francis, and Roger~J Booth. 2001.
\newblock Linguistic inquiry and word count: Liwc 2001.
\newblock \emph{Mahway: Lawrence Erlbaum Associates}, 71(2001):2001.

\bibitem[{Pennington et~al.(2014)Pennington, Socher, and
  Manning}]{pennington2014glove}
Jeffrey Pennington, Richard Socher, and Christopher Manning. 2014.
\newblock Glove: Global vectors for word representation.
\newblock In \emph{Proceedings of the 2014 conference on empirical methods in
  natural language processing (EMNLP)}, pages 1532--1543.

\bibitem[{Rojas-Barahona et~al.(2018)Rojas-Barahona, Tseng, Dai, Mansfield,
  Ramadan, Ultes, Crawford, and Gasic}]{rojas2018deep}
Lina Rojas-Barahona, Bo-Hsiang Tseng, Yinpei Dai, Clare Mansfield, Osman
  Ramadan, Stefan Ultes, Michael Crawford, and Milica Gasic. 2018.
\newblock Deep learning for language understanding of mental health concepts
  derived from cognitive behavioural therapy.
\newblock \emph{arXiv preprint arXiv:1809.00640}.

\bibitem[{Rude et~al.(2004)Rude, Gortner, and Pennebaker}]{rude2004language}
Stephanie Rude, Eva-Maria Gortner, and James Pennebaker. 2004.
\newblock Language use of depressed and depression-vulnerable college students.
\newblock \emph{Cognition \& Emotion}, 18(8):1121--1133.

\bibitem[{Ruder et~al.(2016)Ruder, Ghaffari, and Breslin}]{ruder2016character}
Sebastian Ruder, Parsa Ghaffari, and John~G Breslin. 2016.
\newblock Character-level and multi-channel convolutional neural networks for
  large-scale authorship attribution.
\newblock \emph{arXiv preprint arXiv:1609.06686}.

\bibitem[{Saxena et~al.(2013)Saxena, Funk, and Chisholm}]{saxena2013world}
Shekhar Saxena, Michelle Funk, and Dan Chisholm. 2013.
\newblock World health assembly adopts comprehensive mental health action plan
  2013--2020.
\newblock \emph{The Lancet}, 381(9882):1970--1971.

\bibitem[{Sekuli{\'c} et~al.(2018)Sekuli{\'c}, Gjurkovi{\'c}, and
  {\v{S}}najder}]{sekulic2018not}
Ivan Sekuli{\'c}, Matej Gjurkovi{\'c}, and Jan {\v{S}}najder. 2018.
\newblock Not just depressed: Bipolar disorder prediction on reddit.
\newblock In \emph{9th Workshop on Computational Approaches to Subjectivity,
  Sentiment and Social Media Analysis}.

\bibitem[{Shen and Rudzicz(2017)}]{shen2017detecting}
Judy~Hanwen Shen and Frank Rudzicz. 2017.
\newblock Detecting anxiety through reddit.
\newblock In \emph{Proceedings of the Fourth Workshop on Computational
  Linguistics and Clinical Psychology—From Linguistic Signal to Clinical
  Reality}, pages 58--65.

\bibitem[{Song et~al.(2018)Song, You, Chung, and Park}]{song2018feature}
Hoyun Song, Jinseon You, Jin-Woo Chung, and Jong~C Park. 2018.
\newblock Feature attention network: Interpretable depression detection from
  social media.
\newblock In \emph{Proceedings of the 32nd Pacific Asia Conference on Language,
  Information and Computation}.

\bibitem[{Sun et~al.(2017)Sun, Shrivastava, Singh, and
  Gupta}]{sun2017revisiting}
Chen Sun, Abhinav Shrivastava, Saurabh Singh, and Abhinav Gupta. 2017.
\newblock Revisiting unreasonable effectiveness of data in deep learning era.
\newblock In \emph{Proceedings of the IEEE international conference on computer
  vision}, pages 843--852.

\bibitem[{{\v{S}}uster et~al.(2017){\v{S}}uster, Tulkens, and
  Daelemans}]{vsuster2017short}
Simon {\v{S}}uster, St{\'e}phan Tulkens, and Walter Daelemans. 2017.
\newblock A short review of ethical challenges in clinical natural language
  processing.
\newblock In \emph{Proceedings of the First ACL Workshop on Ethics in Natural
  Language Processing/Hovy, Dirk [edit.]; et al.}, pages 80--87.

\bibitem[{Wykes et~al.(2015)Wykes, Haro, Belli, Obradors-Tarrag{\'o}, Arango,
  Ayuso-Mateos, Bitter, Brunn, Chevreul, Demotes-Mainard
  et~al.}]{wykes2015mental}
Til Wykes, Josep~Maria Haro, Stefano~R Belli, Carla Obradors-Tarrag{\'o}, Celso
  Arango, Jos{\'e}~Luis Ayuso-Mateos, Istv{\'a}n Bitter, Matthias Brunn, Karine
  Chevreul, Jacques Demotes-Mainard, et~al. 2015.
\newblock Mental health research priorities for europe.
\newblock \emph{The Lancet Psychiatry}, 2(11):1036--1042.

\bibitem[{Yang et~al.(2016)Yang, Yang, Dyer, He, Smola, and
  Hovy}]{yang2016hierarchical}
Zichao Yang, Diyi Yang, Chris Dyer, Xiaodong He, Alex Smola, and Eduard Hovy.
  2016.
\newblock Hierarchical attention networks for document classification.
\newblock In \emph{Proceedings of the 2016 conference of the North American
  chapter of the association for computational linguistics: human language
  technologies}, pages 1480--1489.

\bibitem[{Yates et~al.(2017)Yates, Cohan, and Goharian}]{yates2017depression}
Andrew Yates, Arman Cohan, and Nazli Goharian. 2017.
\newblock Depression and self-harm risk assessment in online forums.
\newblock \emph{arXiv preprint arXiv:1709.01848}.

\bibitem[{Zirikly et~al.(2019)Zirikly, Resnik, Uzuner, and
  Hollingshead}]{zirikly-etal-2019-clpsych}
Ayah Zirikly, Philip Resnik, {\"O}zlem Uzuner, and Kristy Hollingshead. 2019.
\newblock {CLP}sych 2019 shared task: Predicting the degree of suicide risk in
  {R}eddit posts.
\newblock In \emph{Proceedings of the Sixth Workshop on Computational
  Linguistics and Clinical Psychology}, pages 24--33, Minneapolis, Minnesota.
  Association for Computational Linguistics.

\bibitem[{Zomick et~al.(2019)Zomick, Levitan, and
  Serper}]{zomick2019linguistic}
Jonathan Zomick, Sarah~Ita Levitan, and Mark Serper. 2019.
\newblock Linguistic analysis of schizophrenia in reddit posts.
\newblock In \emph{Proceedings of the Sixth Workshop on Computational
  Linguistics and Clinical Psychology}, pages 74--83.

\end{thebibliography}

\bibliographystyle{acl_natbib_nourl}

\end{document}